%% file: main.tex
\documentclass[11pt,a4paper,dvipsnames]{article}
\usepackage[hyperref]{acl2020}
\usepackage{times}
\usepackage{latexsym}
\usepackage[pdftex]{graphicx}
\usepackage{adjustbox}
\usepackage{dialogue}
\usepackage{xspace}
\usepackage[ruled]{algorithm2e}
\usepackage{makecell}
\usepackage{array,multirow,graphicx}
\usepackage{soul}
\usepackage{float}
\setul{0.5ex}{0.3ex}

\setcellgapes{4pt}
\usepackage{xspace}
\usepackage[T1]{fontenc}

\usepackage[english]{babel}
\usepackage[utf8x]{inputenc}
\usepackage{mdwlist}

\usepackage{color}

\definecolor{skyblue}{RGB}{194,237,255}
\definecolor{orange}{RGB}{255,204,145}

\newcommand\ya{\textit{yes-and}\xspace}
\newcommand\corpus{{\fontfamily{qtm}\selectfont\textbf{SPOLIN}}\xspace}
\newcommand\corpusfull{\textit{Selected Pairs Of Learnable ImprovisatioN}\xspace}
\newcommand{\datastyle}[1]{\textit{#1}}
\newcommand\spon{\datastyle{Spontaneanation}\xspace}
\newcommand\corn{\datastyle{Cornell}\xspace}
\newcommand\subtle{\datastyle{SubTle}\xspace}
\newcommand\dd{\datastyle{DailyDialog}\xspace}
\newcommand\persona{\datastyle{Persona-chat}\xspace}

\usepackage{microtype}

\aclfinalcopy 
\def\aclpaperid{1193} 

\title{Grounding Conversations with Improvised Dialogues}

\author{Hyundong Cho \and Jonathan May \\
        Information Sciences Institute \\ 
        University of Southern California\\
        \texttt{\{jcho, jonmay\}@isi.edu}}

\begin{document}
\maketitle
\begin{abstract}

Effective dialogue involves grounding, the process of establishing mutual knowledge that is essential for communication between people. Modern dialogue systems are not explicitly trained to build common ground, and therefore overlook this important aspect of communication. Improvisational theater (improv) intrinsically contains a high proportion of dialogue focused on building common ground,  
and makes use of the \ya principle, a strong grounding speech act, to establish coherence and an actionable objective reality. We collect a corpus of more than 26,000 \ya turns, transcribing them from improv dialogues and extracting them from larger, but more sparsely populated movie script dialogue corpora, via a bootstrapped classifier. We fine-tune chit-chat dialogue systems with our corpus to encourage more grounded, relevant conversation and confirm these findings with human evaluations.

\end{abstract}

\input{sections/intro}

\input{sections/data_collection}

\input{sections/data_analysis}

\input{sections/experiments}

\section{Extracting from Other Corpora}
\label{sec:extension}

The approach to classifier-based mining we describe in Section~\ref{sec:cornell} can naturally be applied to other dialogue corpora. We thus next consider mining the gigantic (441M sentence) OpenSubtitles  \cite{Lison2016OpenSubtitles2016EL} collection. As OpenSubtitles contains undesirable material, such as subtitles for media with minimal dialogue, we instead mine from the (3.3M sentence) \subtle corpus \cite{ameixa2013subtitles}, a preprocessed subset of OpenSubtitles that heuristically combines subtitle sequences into dialogue form.

By iterating through half of this corpus, we collect more than 40,000 {\ya}s from it alone, which, when added to \corpus, yields what we call \corpus-extended, which contains about 68,000 {\ya}s, more than 2.5 times the size of the core \corpus. 
Heuristics for finding alternations mean that SubTle's utterances are shorter than those in \spon and \corn, so once the proportion of utterances longer than the average length of  in \spon and \corn (18.5 words) is less than 40\%, we stop further collection in the remainder of the dataset. 
\corpus-extended is available in the same public repository as \corpus. Details of the iterative process as applied to \subtle are in the appendix.

\input{sections/related_work}

\section{Conclusion}

Inspired by {\ya}s in improv, we carefully construct \corpus, a collection of dialogue pairs with responses that are not only coherent with dialogue context but also initiate the next relevant contribution. We extract high-quality {\ya}s from \spon and build a classifier with them, which is then used to mine additional {\ya}s from the \textit{Cornell Movie-Dialogs Corpus}. We further use our mining technique to elicit a corpus of more than 68,000 {\ya} turn pairs, easily the largest collection of this dialogue act known to exist. From human evaluations of dialogue models trained with various data configurations we find that {\corpus} is useful---when including it we are able to build models that can  generate {\ya}s more consistently than when we leave it out. Nevertheless, our models are still inferior at producing good {\ya}s when compared to professional improvisers. We plan to continue our data-driven approach for grounded conversations by expanding our dataset through our iterative data collection process with other larger text-based open-domain dialogue corpora and extend our work to model and collect longer conversations exhibiting more complex improv-backed turns.

\section*{Acknowledgments}

Many thanks to Nanyun Peng and Xinyu Wang for key contributions in a preliminary study, to Paul F. Tompkins, Colin Anderson, and Earwolf for allowing us to include {\ya}s extracted from \spon in \corpus, to Paul Elsberg, Risa Harms, P.T. McNiff, and Peter Schell for initial inspiration, and to Jordan Boyd-Graber for feedback on the final draft. This material is based on research sponsored by the AFRL and DARPA under agreement number FA8650-18-C-7878. The views and conclusions contained herein are those of the authors and should not be interpreted as necessarily representing the official policies or endorsements, either expressed or implied, of the AFRL, DARPA, or the U.S. Government.

\bibliography{references}
\bibliographystyle{acl_natbib}

\appendix

\input{sections/appendix.tex}

\end{document}

%% file: sections/intro.tex
\section{Introduction}

\begin{figure}[h]
    \centering
    \includegraphics[width=3in]{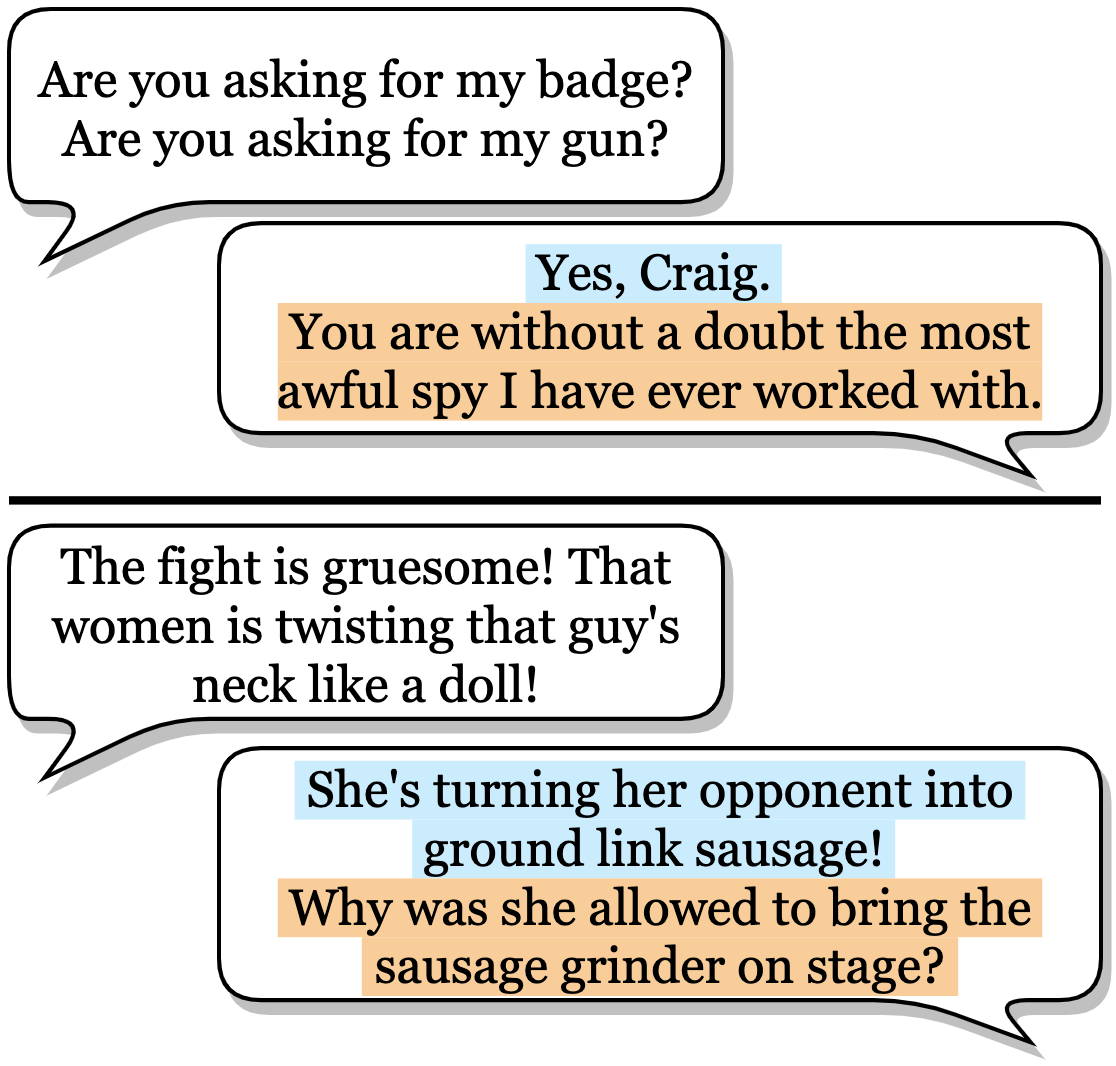}
    \caption{Explicit (top) and implicit (bottom) examples of {\ya}s in the {\corpus} corpus. The text highlighted in \colorbox{skyblue}{light blue} reflects acceptance of the context established in the prompt (``yes'') and the text highlighted in \colorbox{orange}{orange} initiates a new relevant contribution to the dialogue (``and'').}
    \label{fig:corpus_example}
\end{figure}

For humans, dialogue is fundamentally a collaborative, cooperative process by which \textit{partners} coordinate via \textit{turns} or \textit{acts} to jointly construct a common \textit{world state} \cite{bohm2004dialogue}. Without coordination, partners may establish different or conflicting world states, leading to solipsism in the best case and conflict in the worst.   \newcite{clark1989contributing}, describe five dimensions of \textit{grounding}, by which  partners cooperate to establish \textit{common ground}, or a shared world state. The dimension of ``initiation of next relevant contribution'' is the most effective of these in expressing understanding of an ongoing dialogue, and yet is the least observed in  dialogue systems. 

\textit{Improvisational theater} (improv) is a form of theater in which most or all of what is performed is unscripted, created spontaneously by the actors in real time. Because the performance is not scripted and there is typically little to no scenery or other established environment,\footnote{except for, on occasion, external stimulus such as a suggestion from the audience} there is no objective reality that can naturally ground the scene. Hence, actors must mainly rely on dialogue in order to build a coherent scene and progressively establish a common world view. This necessitates accelerated use of the ``initiation of next relevant contribution,'' which in improv is known as the \ya principle. The \ya principle is a rule-of-thumb that suggests that a participant should accept the reality of what the other participant has said (``yes'') and expand or refine that reality with additional information (``and''). Since actors consciously abide by this principle during improv performances, there is a high proportion of these turns embedded in improv dialogue, which helps ensure scenes are coherent and interesting.

Open-domain neural dialogue systems, by contrast, specifically lack coherence and interestingness. 
They commonly repeat previous utterances \cite{jiwei_reinforcement} or generate non-committal, generic statements such as \textit{I don't know} that are logically coherent as a response but preempt further conversation \cite{ sordoni_context_sensitive_generation, serban_hierarchical_generative_nn, li-etal-2016-diversity}.  Either of these developments leads to a conversational black hole and discourages participation in further dialogue turns. This is a critical shortcoming for open-domain dialogue agents, which, unlike task-oriented dialogue systems, are not guided by specific objectives other than entertainment \cite{challenges_in_open_domain_dialogue}. It would behoove such systems to adopt the strategies improvisers include by habit in their dialogues and, consequently, incorporating improv acts should be a key focus for the dialogue community.

Yet, to the best of our knowledge, this has not been previously done. 
There has been work in applying improv to build believable agents that interact with humans \cite{bruce_knight_listopad_magerko_nourbakhsh_2000, winston_magerko} or generate improvised stories \cite{riedl_improv_storytelling}, but development of improv-capable systems in the neural era is largely absent, stymied, we suspect, by the lack of substantial corpora. This is unsurprising; while improv speech acts such as \ya are crucial in all dialogues, they are only highly concentrated in improv dialogues. And improv dialogues are quite difficult to collect; research collections \cite{busso2008scripted} have been far too small to be useful in the modern ML era. The art form has historically been mostly ephemeral, performed live in regional venues on shoestring budgets and rarely recorded.\footnote{The art form has long roots, extending to the Italian \textit{Commedia dell'arte} tradition from the 16th century and farces from the Roman era, but we constrain our focus to the post-20th century form developed and championed by e.g. Keith Johnstone \cite{johnstone2017impro}, Del Close \cite{halpern1994truth}, and our corpus' namesake, Viola Spolin \cite{spolin1986theater}. Spolin was the originator of \textit{Theater Games}, acting exercises that encourage the development of specific theatrical skills. As our corpus is similarly designed to elicit specific skills, we backronym it in recognition of her influence.}   Transcripts are all but absent and mainstream media products are rare.\footnote{One exception, the long-running TV show \textit{Whose Line Is It Anyway}, has, despite a large number of episodes, surprisingly little continuous improvised dialogue, due to the rapid-fire nature of the program.} However, the liberalization of high quality audio \textit{podcasts} 
since 2014 has enabled the availability of a long tail of niche products, improv included \cite{mchugh2016podcasting}.

Therefore we set our objective as collecting \ya-type dialogue pairs ({\ya}s) to enable their modeling by corpus-driven dialogue systems. We mine podcasts and existing movie script corpora for dialogue that abides by the \ya principle and extract
dialogue pairs from these sources to build the {\corpusfull} ({\corpus}) corpus. 
{\corpus} is a collection of more than 26,000 English dialogue turn pairs, each consisting of a \textit{prompt} and subsequent \textit{response}, which abide by the \ya principle, though in diverse manners. Examples of \ya type dialogue pairs collected for {\corpus} are in Figure~\ref{fig:corpus_example}. The corpus is substantial enough to be usable for fine-tuning existing dialogue models to encourage more \ya behavior, and beyond that may prove a valuable knowledge base for empirical sociolinguistic studies on this dialogue act.  

Our contributions are summarized as follows: 

\begin{itemize}
    \item We carefully curate {\corpusfull} ({\corpus}), the first large-scale corpus of \ya dialogue acts, sourced from improv and movie dialogues.
    \item We iteratively build a high-precision \ya classifier, which we use to mine additional {\ya}s from dialogue corpora with high volume but low \ya density. 
    \item We fine-tune existing open-domain conversational models with our corpus and confirm via human evaluations that this approach improves creative grounding. 
    \item We release our models and data for public use, including a 64,000 turn pair extension of the core \corpus, at  \url{https://justin-cho.com/spolin}.
\end{itemize}

%% file: sections/data_collection.tex
\section{Data Collection}
\label{sec:datacollection}

\begin{figure*}[h]
    \centering
    \includegraphics[width=6in]{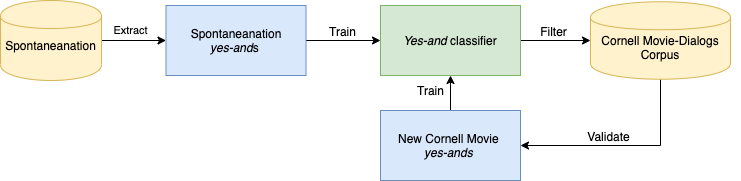}
    \caption{An illustration of the {\ya} collection workflow. The core {\corpus} corpus comprises {\spon} {\ya}s and \corn \xspace{\ya}s (in blue boxes). However, {\corpus} can be augmented by including other general-purpose dialogue corpora in place of \corn in this workflow, as described in Section~\ref{sec:extension}.}
    \label{fig:work-flow}
\end{figure*}

\begin{figure*}
    \centering
    \includegraphics[width=0.75\paperwidth]{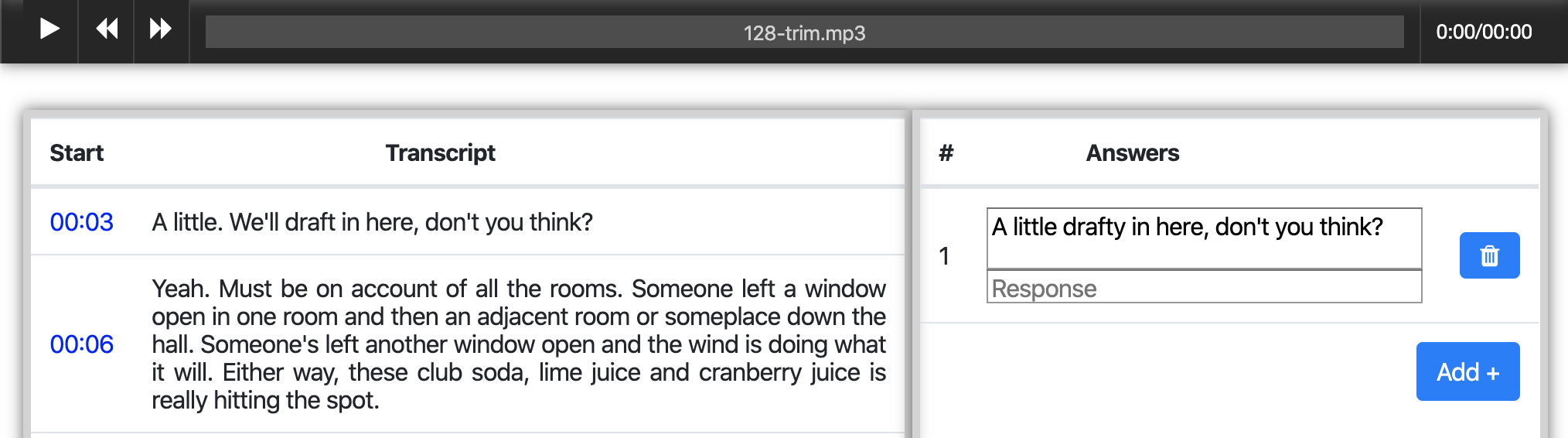}
    \caption{Amazon Mechanical Turk interface for transcribing {\ya}s from {\spon} episodes. Approximate transcriptions with speaker turns and time stamps generated from Amazon Transcribe are provided for additional guidance.}
    \label{fig:spont_interface}
\end{figure*}

\begin{table}[h]
\begin{adjustbox}{width=\columnwidth,center}
    \centering
    \begin{tabular}{lrrrr} 
        Iteration &  1 & 2 & 3 & 4\\ \hline
        {\spon} + & 10,459 &  10,459 & 10,459 & 10,459\\
        {\spon} -- & - & - & 3,225 &5,587 \\
        \corn + & - & 3,327 & 8,464 &12,220\\
        {\corn} -- & 10,459 & 13,786 & 15,698 & 17,092\\
        Total Training Samples & 20,198 & 27,572 & 37,846 & 45,358\\ \hline
        Dev Set Acc. (Spont)& 80.9\%& 73.6\% & 71.6\% &73.0\%\\
        Dev Set Acc. (\corn) & 52.2\%& 56.8\% & 62.1\% & 64.5\%\\ \hline
        Confidence Threshold & 95\% & 70\% & 50\% & 50\% \\ 
        New Extraction Volume & 12,360 & 12,802 & 5,150 & 3,515\\  
        New Proportion of {\ya}s & 26.9\% & 44.0\% & 72.9\% & 78.4\% \\
    \end{tabular}
\end{adjustbox}
    \caption{Iterative data collection results over \corn. + indicates {\ya}s and -- indicates non-{\ya}s. These counts exclude 500 turns collected from each of \spon and \corn to form the validation set. The New Extraction Volume row indicates the new number of \ya  candidates identified with the given confidence threshold, and the New Proportion of {\ya} row show as a percentage how many of these candidates were indeed evaluated as {\ya}s by Turkers. The proportion of {\ya}s increases after each iteration despite the lower confidence threshold used to filter the new predictions with the updated classifier.}
    \label{tab:iteration-table}
\end{table}

Our data collection has five stages: 

\begin{enumerate}
    \item Manually extract {\ya}s from a rich corpus of improv to obtain an initial set of {\ya}s.\label{step:manual}
    \item Construct a \ya \textit{classifier} from the corpus of collected \ya data and negative examples.\label{step:classifier} 
    \item Use the classifier from step \ref{step:classifier} to automatically extract {\ya} candidates from a much larger but sparser dialogue corpus.\label{step:auto}
    \item If necessary, manually validate candidates before adding them to the \ya corpus. 
    \item Repeat from step~\ref{step:classifier} as needed.
\end{enumerate}

An overview of this process is shown in Figure~\ref{fig:work-flow}.

\subsection{Core \ya Collection from {\spon}}

We select the {\spon}\footnote{\url{https://www.earwolf.com/show/spontaneanation-with-paul-f-tompkins/}} podcast as a source of concentrated {\ya}s for its relatively noise-free recording quality and high-quality volume of broad domain improv dialogue. Each episode of this podcast includes an approximately 30 minute-long improv session performed by professional improvisers. Over its 201 episodes, we identified a total of 43K lines of useful spoken dialogue.

Given the confluence of a lack of objective reality, and uninterrupted multiturn dialogue, the improvisers mostly abide by the \ya principle, and therefore {\spon} is a rich resource for natural, high-quality {\ya}s. As it exists only in audio form, and automatic transcription services are too noisy for high quality annotation use, we ask Amazon Mechanical Turk workers (Turkers) to listen to the improv sessions, view Amazon Transcribe preliminary transcriptions, and re-transcribe all of the {\ya}s that they hear using our transcription interface, shown in Figure~\ref{fig:spont_interface}. The interface is based on oTranscribe, an open-source transcription service. Although the quality of transcriptions is poor, we find that including them assists the Turkers in identifying speaker turns and also understanding parts that are sometimes incomprehensible without helping context. 

\subsubsection{Recruiting Quality Crowdworkers for Difficult Annotation Tasks}

One of the main challenges for the data collection process is to recruit competent Turkers who are able to develop a good understanding of the \ya principle.  We actively recruit potential annotators to our task by inviting denizens of the sub-Reddit TurkerNation, rather than simply inviting workers through Amazon's native task posting interface based on HIT approval rate and total number of HITs approved. Our approach enables more human-level engagement, making it easier to determine Turkers' English fluency and experience with improv. To ensure their competence, Turkers first read \ya guidelines (in the appendix) then demonstrate their level of understanding through qualification Human Intelligence Tasks (HITs), which test whether the candidates can identify if a {\ya} exists in a 30 second audio segment and transcribe it if there is one. s

Even after inviting Turkers for the actual HIT of transcribing {\ya}s, we frequently monitor the quality of the data they collect and provide feedback for incorrectly identified {\ya}s. 
Apart from base pay for each episode they work on, we provide incentives for extracting more {\ya}s. The pay for our HITs averages well above California minimum wage.
 From all of the episodes, we extract 10,959 {\ya}s, indicating about 25\% of the total number of dialogue turns in {\spon} are {\ya}s.

\subsection{Guided Extraction from the \textit{Cornell Movie-Dialogs Corpus}}
\label{sec:cornell}

Although larger than any improv corpus, let alone \ya corpus known to date, we seek to increase our corpus volume from 10,959 turn pairs. The \textit{Cornell Movie-Dialogs Corpus} \cite[\corn]{Danescu-Niculescu-Mizil+Lee:11a} contains 304,713 turns, nearly an order of magnitude more than {{\spon}}, and it is one of the closest in domain to improv among existing dialogue datasets. However, a sample annotation of 300 randomly selected turn pairs by Turkers reveal only 11.1\% of pairs are {\ya}s. We thus use the already-collected {\ya}s to probe \corn for likely candidates, to speed the search process. Recent developments of language models pre-trained on massive text data enable the training of high-accuracy models for down-stream tasks even with a small number of samples, by leveraging the contextualized embeddings that these models learn \cite{bert,radford2019language}. We thus fine-tune an initial BERT-based sequence classifier based on the implementation of \newcite{Wolf2019HuggingFacesTS} with the {\ya}s from the {{\spon}} episodes to determine if a given dialogue pair is a {\ya}, using a high threshold (initially, a 95\% probability of being \ya) to bias for precision. We ask Turkers to validate the turn pairs identified by the classifier and add the validated pairs to our \ya corpus. This procedure can be iterated.

\begin{figure*}[h!]
    \centering
    \includegraphics[width=0.75\paperwidth]{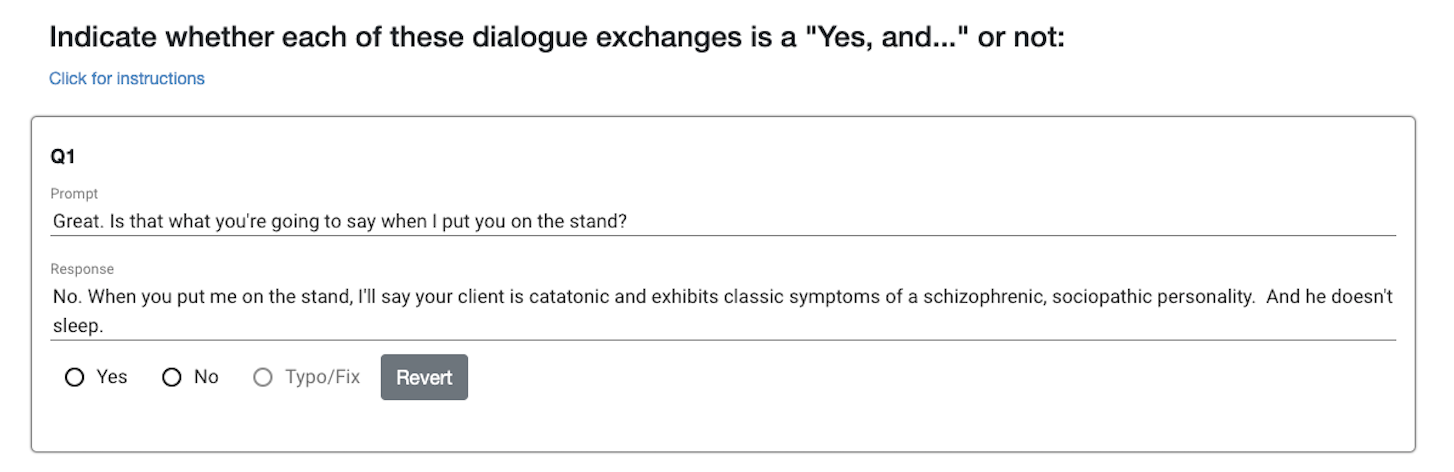}
    \caption{Amazon Mechanical Turk interface for validating {\ya} candidates determined by the {\ya} classifier. Turkers are asked to correct minor errors in grammar, spelling, and punctuation for qualifying {\ya} candidates, which are then categorized as `Typo/Fix.'}
    \label{fig:validation_interface}
\end{figure*}

For the first iteration, we train the classifier with a balanced number of non-{\ya}s chosen by random sampling from \corn, a reasonable assumption due to the relatively low concentration of {\ya}s observed. The same Turkers that extracted {\ya}s from {{\spon}} are invited to validate the {\ya} candidates filtered out by the classifier using the interface shown in Figure~\ref{fig:validation_interface}. In order to ensure consistent annotation standards among Turkers, they are given a small number of overlapping HITs against which we validated. For 90 samples of unfiltered {\ya} candidates from \corn, the two workers yield a reasonably high Cohen's $\kappa$ value of 0.74. Turkers are paid at rates consistent with their rates on the extraction-from-{{\spon}} task.

After the set of \corn \xspace{\ya} candidates are validated, the {\ya}s and non-{\ya}s are added to the training set to train a new classifier, and the same process is repeated. We hold out 500 dialogue pairs from each subcategory (i.e. {\spon}  {\ya}s) as the development set for monitoring the classifier's performance after each iteration. 
We incrementally lower the classification threshold for choosing a {\ya} candidate as the classifier improved. We set this threshold on each iteration except for the first by retrospective evaluation of the classifier on the actual {\ya}  candidates' labels from previous iterations. The threshold with the highest F1 score is chosen to filter new {\ya} candidates to be validated. 

We balance each progressively larger corpus with negative sample turn pairs, which are either randomly selected from \corn (round 1), selected from the rejected-but-extracted turn pairs from \corn (round 2 and later), or sampled from non-\ya turn pairs in \spon formed by random coupling of prompts and responses of the \spon \xspace{\ya}s (round 3 and later). The latter forces the classifier to make decisions based on semantic features relevant to a {\ya} instead of only stylometric features in \spon \xspace {\ya}s. We stop this iterative process after four rounds, when fewer than 5,000 new {\ya} candidates are identified by the classifier, yielding a total corpus size of 26,435 {\ya}s and 23,938 negative samples. An overview of this iterative process is summarized in Table~\ref{tab:iteration-table}. The negative sampling procedure, while somewhat ad-hoc, ultimately provides a mix of turn pairs from both corpora that is sufficient to allow extraction of {\ya}s from new corpora at high precision rates, and is sufficient for our goals.

\begin{table*}[h!]
    \centering
\begin{adjustbox}{max width=\textwidth, center}
    \begin{tabular}{lllr} 
        \multicolumn{2}{c}{Type} & \multicolumn{1}{c}{Example} & \%  \\ \hline
        \parbox[t]{20mm}{\multirow{9}{*}{\ya}} & 
        Explicit  & 
        \makecell[l]{P: Does this map look homemade to you? \\ R: Yeah, it looks like someone without a grasp of English drew it.  }& 15\% \\ {} \\  
        & Implicit & \makecell[l]{P: Alright, pull up that plate so I can take a picture. \\ R: Sorry, the coleslaw is definitely giving off a lot of glare. } & 78\%  \\ {} \\ 
        & \textit{yes-but} & \makecell[l]{ P: We all must say the chant that we say to the king. \\ R: No, it's too erotic, please don't.}  & 7\% \\ {}\\ 
        \parbox[t]{20mm}{\multirow{4}{*}{non-\ya}} & Contra & \makecell[l]{P: Hey, hey, aren't you afraid you'll burn out a tonsil? \\ R: Tonsil? Me? No! Me burn a tonsil?  My tonsils won't burn - As life's corners I...} & 5\% \\ {} \\ 
        & Other & \makecell[l]{P: I feel different right now. \\ R: You wait and see. You're going to marry a big hero!} & 95\% \\ 
    \end{tabular}
\end{adjustbox}
    \caption{Examples and proportions of {\ya} and non-{\ya} types from annotations of 200 {\ya}s and non-{\ya}s in \corpus's development set. Determining whether a given dialogue pair is a {\ya} or not is a non-trivial task, as the agreement or contradiction of the previous dialogue turn's context is usually implicit.}
    \label{tab:type_distribution}
\end{table*}

\subsection{Additional Notes on {\ya} Criteria}

\begin{table}
    \centering
\begin{adjustbox}{max width=\columnwidth, center}
    \begin{tabular}{lrr} 
         & {\ya}s & non-{\ya}s \\ \hline
        \spon & 10,959 & 6,087$^*$ \\
        \corn & 15,476 & 18,351 \\ \hline
        \textbf{Total} & \textbf{26,435} & \textbf{24,438} \\ \hline
    \end{tabular}
\end{adjustbox}
    \caption{Composition of {\corpus}, including the development set. {\ya}s and non-{\ya}s from \corn are validated by Turkers. $^*$\spon non-{\ya}s are sampled from random combination of prompts and responses in {\spon} {\ya}s to balance the dataset for training the classifier in the final iteration, as shown in the last column of Table~\ref{tab:iteration-table}.}
    \label{tab:train_set}
\end{table}

Although the concept of a {\ya} is easy to define and understand, there are borderline cases between a {\ya} and a non-{\ya} that make the validation phase more difficult than originally expected. One of the cases that confused Turkers in the earlier stages of data collection is the case of \textit{yes-but}s. A \textit{yes-but} is a {\ya} with a response that is coherent with the provided reality, but does not appear to provide an affirmative acceptance of a suggestion given in the prompt. These are different from \textit{contradictions} that do not align with the previously established reality. In addition, there are instances where the response is a {\ya}, but is accepted by a speaker other than the one to whom the prompt is directed. 
Some \ya responses initiates a repair of a problem encountered while accepting the prompt, due to a confusion or a possible inconsistency, by asking for clarification \cite{clark1989contributing}. While these responses may not strictly \textit{establish} more detail, they provide information for ultimately establishing new information. We elide these edge cases under the umbrella category \ya in {\corpus} as they further our top-level goal of providing relevant, actionable turn responses. Examples of some of these subtle differences are shown in Table~\ref{tab:type_distribution}.

%% file: sections/data_analysis.tex
\section{Dataset Analysis}

In order to provide a better understanding on the characteristics of our corpus, we annotate 200 {\ya}s and 200 non-{\ya}s in {\corpus}'s development set to categorize them into specific {\ya} or non-{\ya} types. 

We classify {\ya}s into explicit {\ya}s, implicit {\ya}s, or \textit{yes-but}s. Only 15\% of all {\ya}s are explicit {\ya}s, containing phrases such as ``Yeah'' or ``Sure'' that reflects agreement. Even with such phrases, identifying explicit {\ya}s is not a trivial task because it requires semantic understanding of the relevance of the context established in the prompt and that introduced in the response. In fact, there are non-{\ya}s that contain phrases affirming agreement but have no contributions or have new contributions that lack relevance. The majority (78\%) of {\ya}s are implicit {\ya}s, meaning that the agreement is implied, often in a subtle manner. The remaining 7\% are \textit{yes-but}s. 

Non-{\ya}s are divided into \textit{contradictions} and \textit{others}. Most of the non-{\ya} were \textit{other}s, as only 5\% of candidates extracted from \corn are \textit{contradictions}, which are dialogue pairs with a response that actively negates the reality in the prompt. \textit{Other}s encompass any dialogue pairs with a response that lacks coherence to the prompt or adds no or minimal contributions.
The distribution and examples of different types of {\ya}s and non-{\ya}s are shown in Table~\ref{tab:type_distribution}.

The main focus of our work is on {\ya}s, but we provide non-{\ya}s as part of {\corpus} for those interested in training their own classifiers. The negative samples are collected using the methods described in Section~\ref{sec:cornell}. The composition details of {\corpus} are shown in Table~\ref{tab:train_set}. 

%% file: sections/experiments.tex
\section{Experiments}

To evaluate the effect of \corpus on generating \ya responses and thus improving generated dialogue quality, we train a common architecture with a variety of fine-tuning data configurations, both with and without \corpus. Specifically, for each data configuration we fine-tune a doublehead GPT-2 model (117M-parameter version based on the implementation by \newcite{transfer-learning-conversation}), which achieved state-of-the-art performance on \persona for the ConvAI-2 dialogue competition \cite{convai}. We fine-tune the models using two learning objectives, which we weigh equally in calculating loss:

\begin{enumerate}
    \item Predicting the next word.
    \item Predicting the next correct candidate that best fits the dialogue given the dialogue history.
\end{enumerate}

The language modeling component uses pre-trained weights from OpenAI, while the candidate classification head is trained from scratch. For evaluation, we use the language modeling component of the fine-tuned model to generate single-turn responses for the {\ya} prompts in the development set. We use nucleus sampling \cite{holtzman2019curious} for the decoding step to keep only the top tokens with a cumulative probability that together exceed 0.9, from which the next token is chosen with multinomial sampling.

\subsection{Data Configurations}

\begin{figure}[h]
    \centering
    \includegraphics[width=\columnwidth]{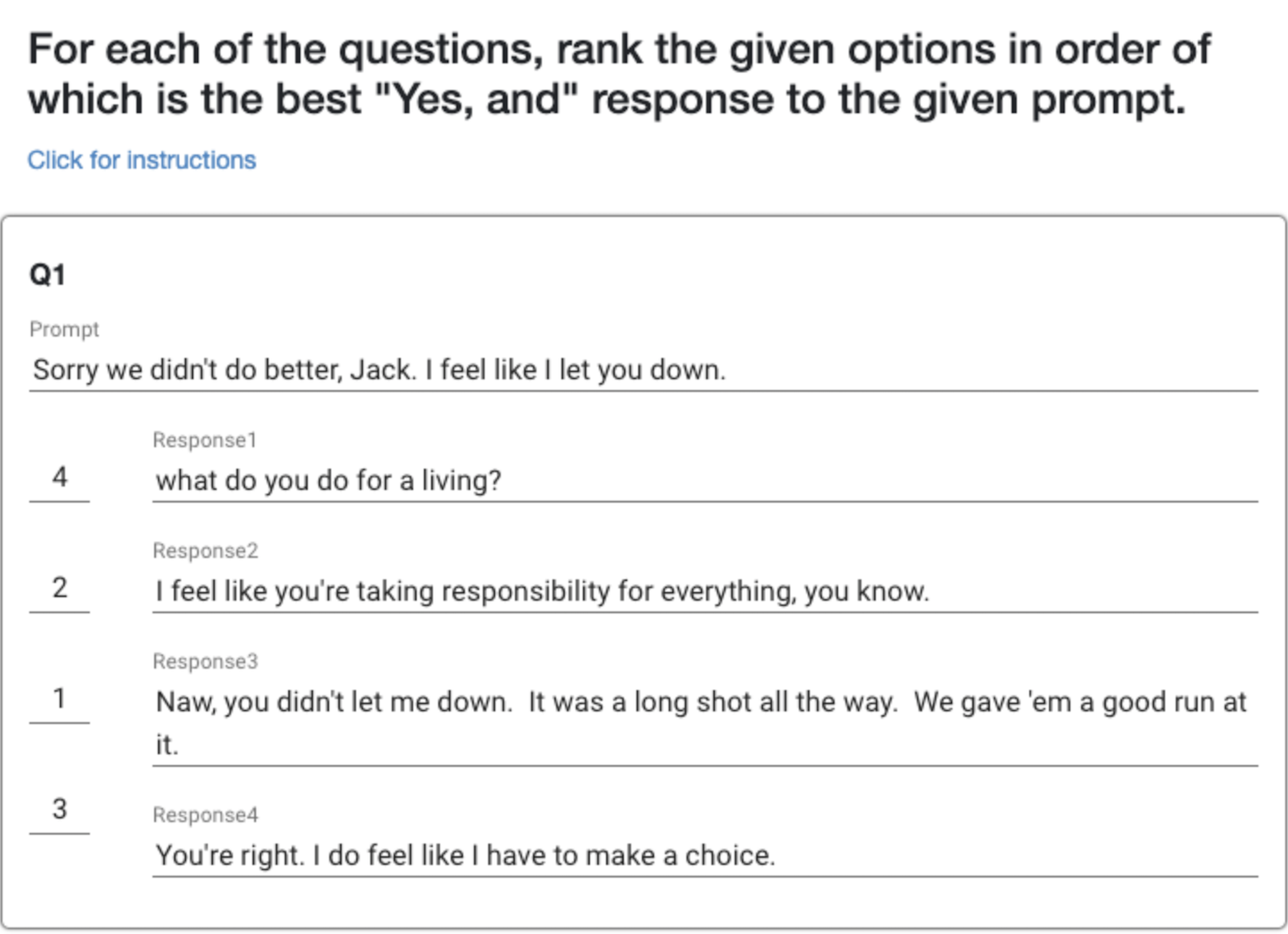}
    \caption{Interface used by human evaluators to rank responses based on their quality as a {\ya}, where a rank of 1 is most preferred. The correct ranking is shown for this example. The option ranked 1 is a \textit{yes-but}: it does not reject a reality but rather rejects a suggestion and provides refining information that is most coherent to the prompt.}
    \label{fig:ranking_interface}
\end{figure}

For our experiments, we use several established dialogue datasets as baselines, namely \persona \cite{convai}, \corn \cite{Danescu-Niculescu-Mizil+Lee:11a} (the unfiltered corpus out of which we extract {\ya}s, as described in Section~\ref{sec:cornell}), and {\dd} \cite{li-etal-2017-dailydialog}. Each of these is an English-language open-domain casual conversation corpus with 100k--300k turns. For each of these datasets, we either simply fine-tune on that dataset, or fine-tune and then further fine-tune with \corpus. In another configuration, we also fine-tune directly with \corpus on top of GPT-2. The original GPT-2 implementation prepends the personalities given in {\persona} to the dialogue sequence input before tokenization. For fine-tuning to datasets apart from {\persona}, we simply do not prepend any auxiliary information to the dialogue sequence input.

\subsection{Human Evaluation}

Automatic metrics that rely on n-gram overlap, such as BLEU, ROUGE, and METEOR, are often used for generative models when there is little variability in the target output \cite{papineni2002bleu, lin-2004-rouge, banerjee2005meteor}. However, there can be a wide variety of responses that qualify as a good {\ya}, a problem common to open-domain generation tasks. An adequate evaluation of our models requires assessing the main {\ya} criteria: agreement with the context and the quality of the new relevant contribution, both of which are not feasible with the aforementioned metrics. Therefore, we ask human evaluators to compare the quality of the {\ya}s generated by various models and the actual response to the prompt in {\corpus} that is used as the input.

\begin{table*}[h]
    \centering
\begin{adjustbox}{max width=\textwidth, center}
    \begin{tabular}{p{40mm}r|p{40mm}|p{90mm}}
        Dataset &  \multicolumn{1}{r}{Avg Rank $\downarrow$}  & \multicolumn{1}{l}{Example Prompt} & Example Responses \\ \Xhline{2\arrayrulewidth} \hline
        \persona& 3.67 & \multirow{5}{40mm}{\texttt{I know alotta women and I'm sure she remembers me.}} & \texttt{oh my goodness, i don't know her} \\  \cline{1-2} \cline{4-4}
        {\corpus}  & 3.41 & & \texttt{Yeah, she's a bit of a mystery.} \\  \cline{1-2} \cline{4-4}
        \multirow{2}{40mm}{{\persona}+{\corpus}} & \multirow{2}{*}{\textbf{3.17}} &  & \texttt{So you remember her? I remember her in the shower}. \\  \cline{1-2} \cline{4-4}
        Gold & 1.92 & &\texttt{She does.  From when you were a boy.} \\  \hline\hline
        \corn& 3.73 & {\multirow{6}{40mm}{\texttt{I made this beautiful salmon mousse that just looked like a puddle of spit.}}} & \texttt{What?} \\  \cline{1-2} \cline{4-4}
        {\corpus}  & 3.39 & & \texttt{And it's delicious!}\\  \cline{1-2} \cline{4-4}
        \multirow{2}{40mm}{{\corn}+{\corpus}} & \multirow{2}{*}{\textbf{3.34}} &  & \texttt{That's the kind of thing you do when you're in love.} \\  \cline{1-2} \cline{4-4}
        \multirow{2}{40mm}{Gold} & \multirow{2}{*}{2.01} & & \texttt{It was genius. It's making me hungry thinking about it.} \\ \hline\hline
        \multirow{2}{40mm}{\dd} & \multirow{2}{*}{3.37}  & \multirow{7}{40mm}{\texttt{Excuse me. Is anybody in here? I'd like a bagel. Hello? Is anybody in here?}} & \texttt{I'm sorry, sir. I'm not in the mood. I'm not in the mood.}  \\  \cline{1-2} \cline{4-4}
        \multirow{2}{40mm}{\corpus}  & \multirow{2}{*}{3.32} & &  \texttt{I'm in here. I'm just trying to make sure I can get a bagel.} \\  \cline{1-2} \cline{4-4}
        {\dd}+{\corpus} & \textbf{3.31} &  & \texttt{Oh, yeah, the guy who left the bagel.} \\  \cline{1-2} \cline{4-4}
        \multirow{2}{40mm}{Gold} & \multirow{2}{*}{1.87} & & \texttt{I can help you. The problem is that the bagels are burned.} \\ \hline\hline
    \end{tabular}
\end{adjustbox}
    \caption{Average human ranking of responses to prompts from \spon generated by models trained with {\corpus}, an existing dialog corpus, or both, based on the {\ya} criteria. Rank is scaled from 1 to 4, lower is better.}
    \label{tab:yesand-rank}
\end{table*}

 \begin{table*}[h!]
 \begin{adjustbox}{width=\textwidth,center}
     \centering
     \begin{tabular}{lll}
        Dataset  & Source & Size$^*$ \\ \hline 
        DailyDialog \cite{li-etal-2017-dailydialog} & Crowdsourced & 104K \\
        Cornell Movie-Dialogs Corpus \cite{Danescu-Niculescu-Mizil+Lee:11a} & Movie scripts &  304K \\
        Persona-chat \cite{convai} & Crowdsourced & 162K \\ 
        The Ubuntu Dialogue Corpus \cite{ubuntu} & Ubuntu chat logs & 7M \\
        Twitter Triple Conversations \cite{sordoni_context_sensitive_generation} & Social media & 6K \\
        OpenSubtitles \cite{Lison2016OpenSubtitles2016EL} & Subtitles & 441M sentences \\ 
        SubTle (Eng) \cite{ameixa2013subtitles} & Subtitles & 3.3M pairs \\ 
        London-Lund Corpus \cite{greenbaum1990london} & Various sources & 500K words \\
        London-Lund Corpus 2 \cite{poldvere2017london} & Various sources & 500K words \\ \hline
        \textbf{\corpus} ({\ya} only) & Improv, Movie scripts &  26K pairs \\ 
        \textbf{\corpus-extended} ({\ya} only) & Improv, Movie scripts, subtitles & 68K pairs \\ \hline 
     \end{tabular}
\end{adjustbox}
     \caption{A survey of publicly available English language text-based corpora frequently used for open-domain dialogue systems. The last two rows are our contribution. $^*$Size is measured as the number of total utterances (dialogue turns) unless otherwise specified.}
     \label{tab:corpora_survey}
 \end{table*}

We ask human evaluators to rank a set of four responses given a prompt, comparing the responses of a model trained only with {\corpus}, a model trained with an existing dialogue corpus, a model trained with both, and the actual response pair from the development set, denoted as ``Gold.'' These four responses are randomly ordered for each question to prevent evaluators from developing a bias for responses that frequently have a good or poor response in a set order, as shown in Figure \ref{fig:ranking_interface}. The evaluators are permitted to provide the same rank for different responses if they are equal in quality. This evaluation set contains 100 such prompts, and each is evaluated twice by different evaluators. The results of the average ranking and some of the examples generated by the models are shown in Table~\ref{tab:yesand-rank}.

Results show that models trained only with {\corpus} or with {\corpus} and another dialogue dataset are preferred to the models trained only with another dialogue dataset, although in the case of \dd, the average ranking improves only by at most 0.06 after fine-tuning with {\corpus}. However, even the responses generated by models trained with {\corpus} are not ranked as well as the actual responses in the development set, indicating our models are still inferior to professional human improviser quality.

%% file: sections/related_work.tex
\section{Related Work}

Many works have identified the same issues of repetitive or non-committal responses generated by neural conversational systems that are at least partially related to the lack of sufficiently high quality {\ya}s we deal with in this work;  approaches that mitigate these problems vary. The majority of recent works focus on diversifying the responses by modifying the training and decoding objectives \cite{li-etal-2016-diversity, jiwei_decode, li-etal-2017-adversarial, jiwei_reinforcement, xu-etal-2017-neural, shao2017}. Other methods introduce latent variables to encourage diversity \cite{serban2017hierarchical, zhao2017diversity_vae}. Some explore methods of re-weighing training instances that encourage diversity \cite{liu-etal-2018-towards-less, lison-bibauw-2017-dialogues, du-black-2019-boosting}. 

Our approach is complementary to all the model-based approaches described here, as it simply deals with the production of a particularly useful \textit{corpus}, that can be used to fine-tune on top of these methods.

We provide a survey of publicly available text-based datasets frequently used for open-domain dialogue systems and discuss their limitations for our purpose of generating grounded responses (see Table \ref{tab:corpora_survey} for an overview).

\textit{DailyDialog} is a collection of multi-turn dialogue with manually annotated emotion and intent labels \cite{li-etal-2017-dailydialog}. \newcite{Danescu-Niculescu-Mizil+Lee:11a} created \textit{the Cornell Movie-Dialogs Corpus}, a compilation of dialogue sequences paired with meta data about the movie and characters. \persona provides dialogue sequence coupled with corresponding personas \cite{convai}. The \textit{Ubuntu Dialogue Corpus} contains 1 million dialogue turns extracted from Ubuntu chat logs, which discuss Ubuntu-related technical support \cite{ubuntu}. The \textit{Twitter Triple Corpus} is a dataset of 4K dialogue triples extracted from Twitter \cite{sordoni_context_sensitive_generation}.  \textit{OpenSubtitles} is a huge collection of subtitles that span various genres, but the absence of speaker turn annotations make it difficult to modify into dialogue format \cite{Lison2016OpenSubtitles2016EL}. \newcite{ameixa2013subtitles} use heuristics to reformat OpenSubtitles into dialogues with some limited success. \newcite{clark1989contributing} illustrate grounding in conversations with examples from the London-Lund Corpus \cite{greenbaum1990london}, a corpus of full conversations annotated with prosodic and paralinguistic features. A second version of the corpus was compiled with the same annotations standards as the first using more recent spoken and text data \cite{poldvere2017london}. 

These corpora were not collected with the criteria for {\ya}s in mind. Even for datasets with dialogue taking place in a similar domain as improv, they naturally contain only a small proportion of {\ya}s. However, the relatively large sizes of these datasets still make them useful for dialogue systems. They can be used effectively for grounded conversations if the {\ya}s or other desirable dialogue acts can be filtered out or given higher weights in training to enforce their characteristics in the responses generated. 

Our data collection approach is similar to the method of \newcite{yarowsky-1995-unsupervised}, which formalizes the bootstrapping mechanism of iteratively improving a classifier and label unlabeled data. 
The main difference from the Yarowsky algorithm and our approach is that, rather than using a fully automated process for increasing training data, we use a probability threshold to regulate recall, followed by human judgment to ensure high precision. 

 
 Apart from \newcite{clark1989contributing} there have been other taxonomies of grounding. For example, \newcite{traum1999computational} considers six categories; among these are \textit{acknowledge} and \textit{continue}, which, taken together, map nicely to \ya. \newcite{magerkoemp} and \newcite{sharedmental} note the importance of establishing common ground in improv. 

%% file: sections/appendix.tex
\section{Appendix}
\label{sec:appendix}

\begin{table}[h!]
\begin{adjustbox}{width=\columnwidth}
    \centering
    \begin{tabular}{lrrrrrrr} 
        Iteration & 4 & 5 & 6 & 7\\ \hline
        {\spon} +  & 10,459 & 10,459 & 10,459 & 10,459 \\
        {\spon} - &5,587 & 5,587 & 5,587 & 5,587  \\
        \corn +  &12,220 & 14,976 & 14,976 & 14,976 \\
        \corn -  & 17,092 & 17,701 & 17,701 & 17,701 \\
        \subtle +  & - &  2,621 & 20,617 & 33,155 \\
        \subtle - & - &  7,865 & 14,799 & 17,325 \\
        Total Training Samples & 45,358 & 59,209 & 84,319 & 99,203 \\ \hline
        Dev Set Acc. (Spont)&73.0\% & 72.1\% & 68.4\% & 75.2\% \\
        Dev Set Acc. (\corn) & 64.5\% & 63.3\% & 63.3\% & 61.0\% \\ \hline
        Confidence Threshold & 50\% / 70\%* & 70\% & 70\% & 70\% \\ 
        New Extraction Volume & 3,515 / 10,486* & 36,608 & 15,424 & 14,979 \\  
        New Proportion of {\ya}s & 78.4\% / 25.0\%*  & 58.4\% & 83.2\% & 76.0\% \\  \hline
    \end{tabular}
\end{adjustbox}
    \caption{Continuation of Table \ref{tab:iteration-table} with the extended version of \corpus that includes extracted {\ya}s from \subtle. \subtle is collected from the fourth iteration onwards.  *Statistics for \corn/\subtle are shown separately. The same classifier is used for extracting candidates from \corn and \subtle, but they are datasets with significantly different characteristics.}
    \label{tab:subtle-table}
\end{table}

\subsection{\ya Guidelines for Turkers}

\begin{figure*}
    \centering
    \includegraphics[width=0.75\paperwidth]{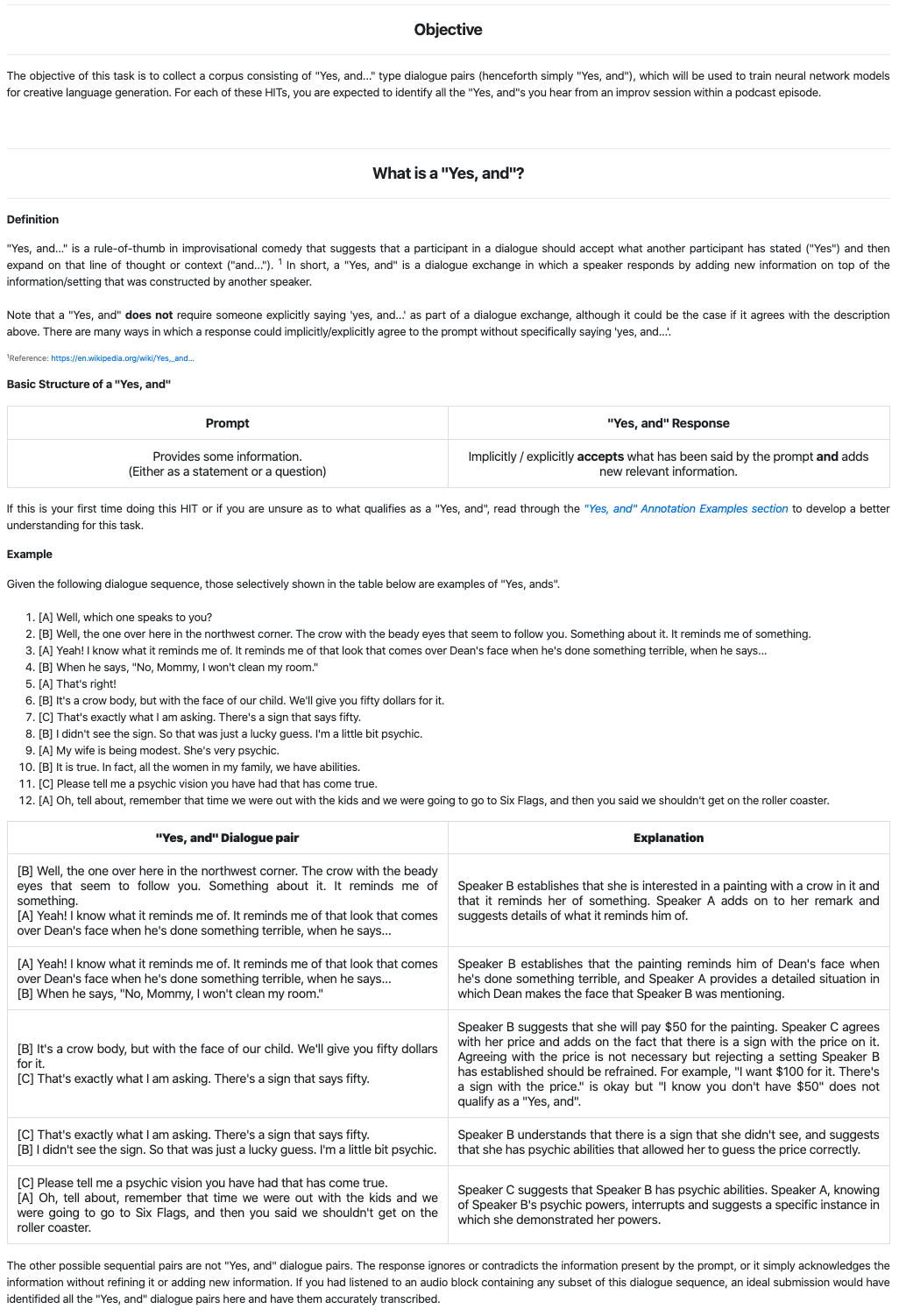}
    \caption{Explanation of the objective and \ya in the \ya guideline provided to Turkers.}
    \label{fig:yesand-guide1}
\end{figure*}

\begin{figure*}
    \centering
    \includegraphics[width=0.75\paperwidth]{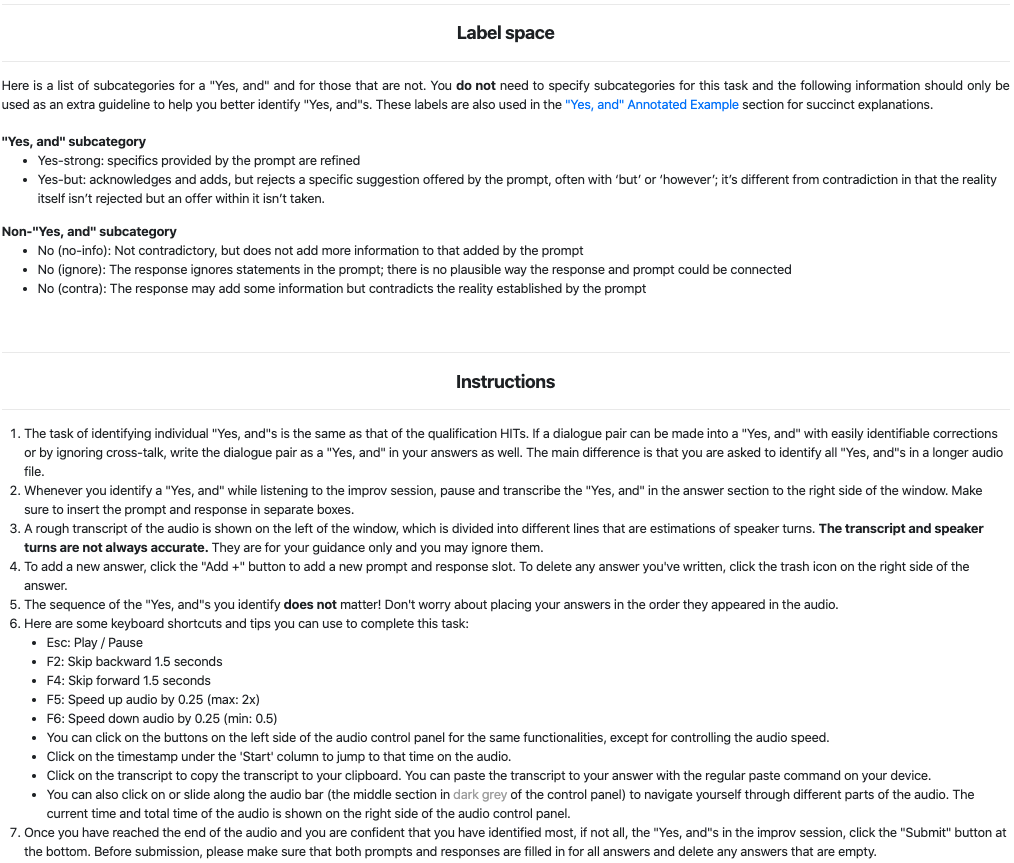}
    \caption{Explanation of the label space for {\ya}s and non-{\ya}s and the detailed instructions for the transcription task.}
    \label{fig:yesand-guide2}
\end{figure*}

\begin{figure*}
    \centering
    \includegraphics[width=0.75\paperwidth]{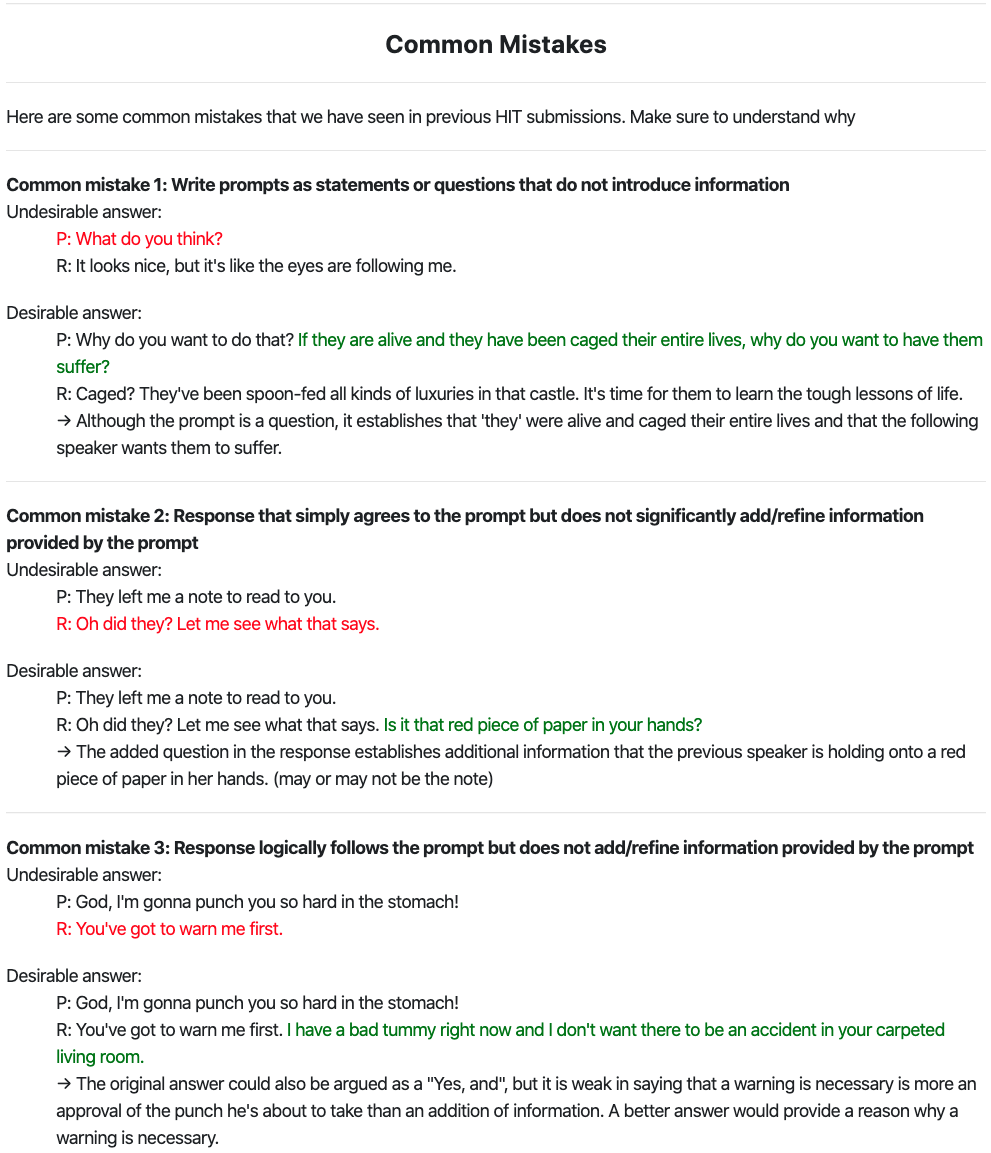}
    \caption{Common mistakes that Turkers made in the early stages of data collection were corrected and added to the guidelines to aid new Turkers.}
    \label{fig:common-mistakes}
\end{figure*}

\begin{figure*}
    \centering
    \includegraphics[width=0.75\paperwidth]{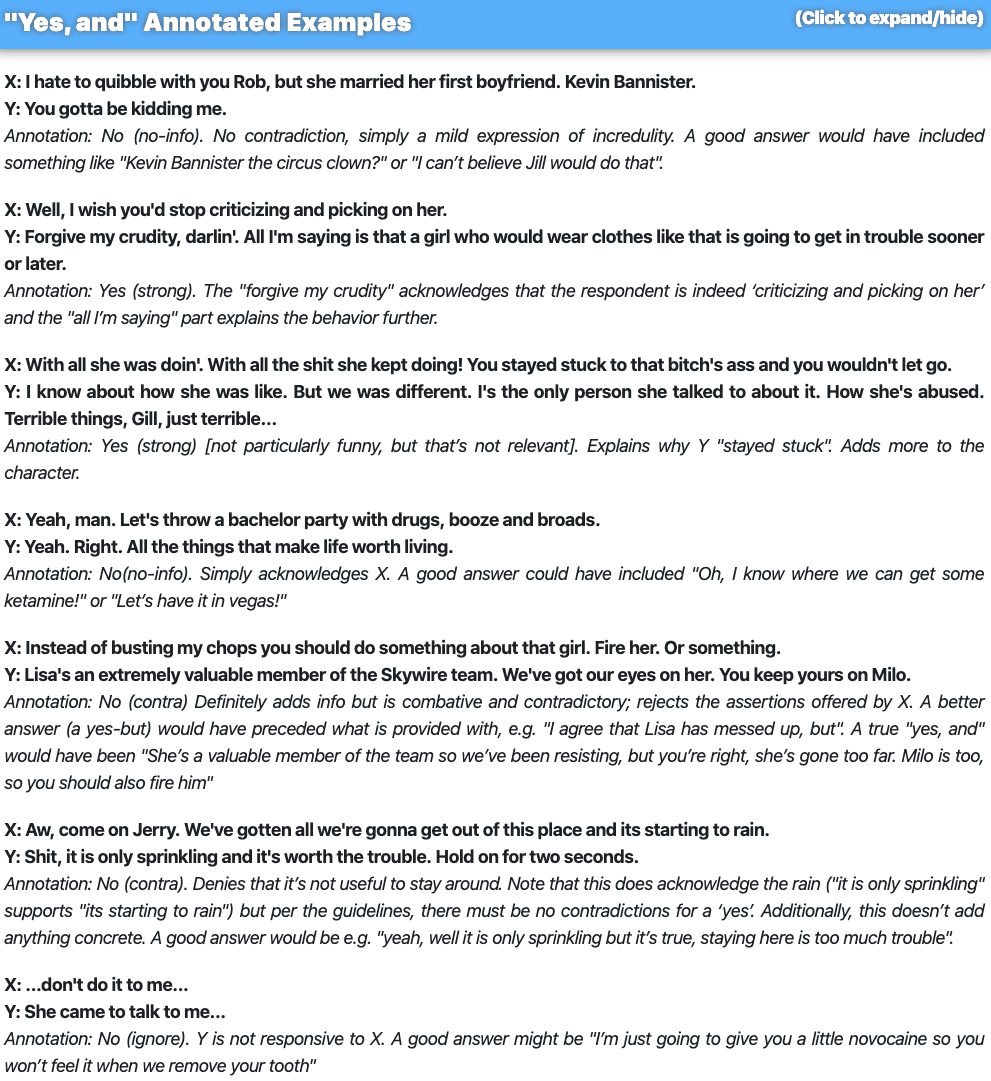}
    \caption{Annotated examples provided to Turkers for understanding the label space of the \ya transcription task.}
    \label{fig:annotation-examples}
\end{figure*}

We provide detailed annotation guidelines, shown in Figures \ref{fig:yesand-guide1}--\ref{fig:annotation-examples}, to the Turkers as a result of having continuous discussions with them and monitoring their submissions. Contrary to our expectations, it is difficult to make a binary decision on whether a dialogue turn is a \ya or non-\ya, and therefore these fine-grained details are crucial for collecting {\ya}s in \corpus.

\subsection{Iterative data collection results for \subtle}

Due to \subtle's relatively large size, we split \subtle into 20 equal blocks that each contains 10,486 dialogue turns. Note that every successive iteration of \subtle was not performed on the same block but on the next block. This is different from \corn, for which every iteration is on the same set of dialogue turns. This difference is not due to any characteristics in the dataset but because of practical reasons arising from the size of the \subtle corpus. 

The first extraction proportion for \subtle is low because of the prevalence of self-{\ya}s in this corpus. Self-{\ya}s are prompt and response pairs that evidently originate from the same speaker but otherwise meet the criteria of a {\ya}. There are many incorrectly combined dialogue turns that actually come from the same speaker because of the heuristics employed for building \subtle. By providing labeled self-{\ya} as negative samples, the classifier quickly learns to remove these self-{\ya}s, leading to a significantly higher proportion of {\ya}s in subsequent iterations. This is demonstrated in the specifics of the additional iterations, which are shown in Table \ref{tab:subtle-table}.

%% file: main.bbl
\begin{thebibliography}{44}
\expandafter\ifx\csname natexlab\endcsname\relax\def\natexlab#1{#1}\fi

\bibitem[{Ameixa et~al.(2013)Ameixa, Coheur, and Redol}]{ameixa2013subtitles}
David Ameixa, Lu{\'\i}sa Coheur, and Rua~Alves Redol. 2013.
\newblock From subtitles to human interactions: {I}ntroducing the {SubTle}
  corpus.
\newblock Technical report, INESC-ID.

\bibitem[{Banerjee and Lavie(2005)}]{banerjee2005meteor}
Satanjeev Banerjee and Alon Lavie. 2005.
\newblock \href {https://www.aclweb.org/anthology/W05-0909} {{METEOR}: An
  automatic metric for {MT} evaluation with improved correlation with human
  judgments}.
\newblock In \emph{Proceedings of the {ACL} Workshop on Intrinsic and Extrinsic
  Evaluation Measures for Machine Translation and/or Summarization}, pages
  65--72, Ann Arbor, Michigan. Association for Computational Linguistics.

\bibitem[{Bohm and Nichol(2004)}]{bohm2004dialogue}
David Bohm and Lee Nichol. 2004.
\newblock \href {https://books.google.com/books?id=MGGF\_oF\_aY0C} {\emph{On
  Dialogue}}.
\newblock Routledge classics. Routledge.

\bibitem[{Bruce et~al.(2000)Bruce, Knight, Listopad, Magerko, and
  Nourbakhsh}]{bruce_knight_listopad_magerko_nourbakhsh_2000}
Allison Bruce, Jonathan Knight, Samuel Listopad, Brian Magerko, and Illah~R.
  Nourbakhsh. 2000.
\newblock \href {https://doi.org/10.1109/robot.2000.845355} {Robot improv:
  {u}sing drama to create believable agents}.
\newblock \emph{Proceedings 2000 ICRA. Millennium Conference. IEEE
  International Conference on Robotics and Automation. Symposia Proceedings
  (Cat. No.00CH37065)}.

\bibitem[{Busso and Narayanan(2008)}]{busso2008scripted}
Carlos Busso and Shrikanth~S Narayanan. 2008.
\newblock Scripted dialogs versus improvisation: Lessons learned about
  emotional elicitation techniques from the {IEMOCAP} database.
\newblock In \emph{Ninth annual conference of the international speech
  communication association}.

\bibitem[{Clark and Schaefer(1989)}]{clark1989contributing}
Herbert~H Clark and Edward~F Schaefer. 1989.
\newblock Contributing to discourse.
\newblock \emph{Cognitive science}, 13(2):259--294.

\bibitem[{Danescu-Niculescu-Mizil and
  Lee(2011)}]{Danescu-Niculescu-Mizil+Lee:11a}
Cristian Danescu-Niculescu-Mizil and Lillian Lee. 2011.
\newblock \href {https://www.aclweb.org/anthology/W11-0609} {Chameleons in
  imagined conversations: A new approach to understanding coordination of
  linguistic style in dialogs}.
\newblock In \emph{Proceedings of the 2nd Workshop on Cognitive Modeling and
  Computational Linguistics}, pages 76--87, Portland, Oregon, USA. Association
  for Computational Linguistics.

\bibitem[{Devlin et~al.(2019)Devlin, Chang, Lee, and Toutanova}]{bert}
Jacob Devlin, Ming-Wei Chang, Kenton Lee, and Kristina Toutanova. 2019.
\newblock \href {https://doi.org/10.18653/v1/N19-1423} {{BERT}: Pre-training of
  deep bidirectional transformers for language understanding}.
\newblock In \emph{Proceedings of the 2019 Conference of the North {A}merican
  Chapter of the Association for Computational Linguistics: Human Language
  Technologies, Volume 1 (Long and Short Papers)}, pages 4171--4186,
  Minneapolis, Minnesota. Association for Computational Linguistics.

\bibitem[{Du and Black(2019)}]{du-black-2019-boosting}
Wenchao Du and Alan~W Black. 2019.
\newblock \href {https://doi.org/10.18653/v1/P19-1005} {Boosting dialog
  response generation}.
\newblock In \emph{Proceedings of the 57th Annual Meeting of the Association
  for Computational Linguistics}, pages 38--43, Florence, Italy. Association
  for Computational Linguistics.

\bibitem[{Fuller and Magerko(2010)}]{sharedmental}
Daniel Fuller and Brian Magerko. 2010.
\newblock \href {https://doi.org/10.1145/1822309.1822324} {Shared mental models
  in improvisational performance}.
\newblock In \emph{Proceedings of the Intelligent Narrative Technologies III
  Workshop}, INT3 ’10, New York, NY, USA. Association for Computing
  Machinery.

\bibitem[{Greenbaum and Svartvik(1990)}]{greenbaum1990london}
Sidney Greenbaum and Jan Svartvik. 1990.
\newblock \emph{The {L}ondon-{L}und corpus of spoken {E}nglish}, volume~7.
\newblock Lund University Press.

\bibitem[{Halpern et~al.(1994)Halpern, Close, and Johnson}]{halpern1994truth}
Charna Halpern, Del Close, and Kim Johnson. 1994.
\newblock \emph{Truth in comedy: The manual of improvisation}.
\newblock Meriwether Publishing.

\bibitem[{Holtzman et~al.(2020)Holtzman, Buys, Forbes, and
  Choi}]{holtzman2019curious}
Ari Holtzman, Jan Buys, Maxwell Forbes, and Yejin Choi. 2020.
\newblock The curious case of neural text degeneration.
\newblock In \emph{Proceedings of the Eighth International Conference on
  Learning Representations}.

\bibitem[{Huang et~al.(2020)Huang, Zhu, and
  Gao}]{challenges_in_open_domain_dialogue}
Minlie Huang, Xiaoyan Zhu, and Jianfeng Gao. 2020.
\newblock \href {https://doi.org/10.1145/3383123} {Challenges in building
  intelligent open-domain dialog systems}.
\newblock \emph{ACM Transactions on Information Systems}, 38(3):1–32.

\bibitem[{Johnstone(2017)}]{johnstone2017impro}
Keith Johnstone. 2017.
\newblock \href {https://books.google.com/books?id=j0n2DAAAQBAJ} {\emph{Impro:
  Improvisation and the Theatre}}.
\newblock Performance Books. Bloomsbury Publishing.

\bibitem[{Li et~al.(2016{\natexlab{a}})Li, Galley, Brockett, Gao, and
  Dolan}]{li-etal-2016-diversity}
Jiwei Li, Michel Galley, Chris Brockett, Jianfeng Gao, and Bill Dolan.
  2016{\natexlab{a}}.
\newblock \href {https://doi.org/10.18653/v1/N16-1014} {A diversity-promoting
  objective function for neural conversation models}.
\newblock In \emph{Proceedings of the 2016 Conference of the North {A}merican
  Chapter of the Association for Computational Linguistics: Human Language
  Technologies}, pages 110--119, San Diego, California. Association for
  Computational Linguistics.

\bibitem[{Li et~al.(2016{\natexlab{b}})Li, Monroe, and Jurafsky}]{jiwei_decode}
Jiwei Li, Will Monroe, and Dan Jurafsky. 2016{\natexlab{b}}.
\newblock A simple, fast diverse decoding algorithm for neural generation.
\newblock \emph{arXiv preprint arXiv:1611.08562}.

\bibitem[{Li et~al.(2016{\natexlab{c}})Li, Monroe, Ritter, Jurafsky, Galley,
  and Gao}]{jiwei_reinforcement}
Jiwei Li, Will Monroe, Alan Ritter, Dan Jurafsky, Michel Galley, and Jianfeng
  Gao. 2016{\natexlab{c}}.
\newblock \href {https://doi.org/10.18653/v1/D16-1127} {Deep reinforcement
  learning for dialogue generation}.
\newblock In \emph{Proceedings of the 2016 Conference on Empirical Methods in
  Natural Language Processing}, pages 1192--1202, Austin, Texas. Association
  for Computational Linguistics.

\bibitem[{Li et~al.(2017{\natexlab{a}})Li, Monroe, Shi, Jean, Ritter, and
  Jurafsky}]{li-etal-2017-adversarial}
Jiwei Li, Will Monroe, Tianlin Shi, S{\'e}bastien Jean, Alan Ritter, and Dan
  Jurafsky. 2017{\natexlab{a}}.
\newblock \href {https://doi.org/10.18653/v1/D17-1230} {Adversarial learning
  for neural dialogue generation}.
\newblock In \emph{Proceedings of the 2017 Conference on Empirical Methods in
  Natural Language Processing}, pages 2157--2169, Copenhagen, Denmark.
  Association for Computational Linguistics.

\bibitem[{Li et~al.(2017{\natexlab{b}})Li, Su, Shen, Li, Cao, and
  Niu}]{li-etal-2017-dailydialog}
Yanran Li, Hui Su, Xiaoyu Shen, Wenjie Li, Ziqiang Cao, and Shuzi Niu.
  2017{\natexlab{b}}.
\newblock \href {https://www.aclweb.org/anthology/I17-1099} {{D}aily{D}ialog: A
  manually labelled multi-turn dialogue dataset}.
\newblock In \emph{Proceedings of the Eighth International Joint Conference on
  Natural Language Processing (Volume 1: Long Papers)}, pages 986--995, Taipei,
  Taiwan. Asian Federation of Natural Language Processing.

\bibitem[{Lin(2004)}]{lin-2004-rouge}
Chin-Yew Lin. 2004.
\newblock \href {https://www.aclweb.org/anthology/W04-1013} {{ROUGE}: A package
  for automatic evaluation of summaries}.
\newblock In \emph{Text Summarization Branches Out}, pages 74--81, Barcelona,
  Spain. Association for Computational Linguistics.

\bibitem[{Lison and Bibauw(2017)}]{lison-bibauw-2017-dialogues}
Pierre Lison and Serge Bibauw. 2017.
\newblock \href {https://doi.org/10.18653/v1/W17-5546} {Not all dialogues are
  created equal: {I}nstance weighting for neural conversational models}.
\newblock In \emph{Proceedings of the 18th Annual {SIG}dial Meeting on
  Discourse and Dialogue}, pages 384--394, Saarbr{\"u}cken, Germany.
  Association for Computational Linguistics.

\bibitem[{Lison and Tiedemann(2016)}]{Lison2016OpenSubtitles2016EL}
Pierre Lison and J{\"o}rg Tiedemann. 2016.
\newblock Open{S}ubtitles2016: Extracting large parallel corpora from movie and
  {TV} subtitles.
\newblock In \emph{LREC}.

\bibitem[{Liu et~al.(2018)Liu, Bi, Gao, Liu, Yao, and
  Shi}]{liu-etal-2018-towards-less}
Yahui Liu, Wei Bi, Jun Gao, Xiaojiang Liu, Jian Yao, and Shuming Shi. 2018.
\newblock \href {https://doi.org/10.18653/v1/D18-1297} {Towards less generic
  responses in neural conversation models: {A} statistical re-weighting
  method}.
\newblock In \emph{Proceedings of the 2018 Conference on Empirical Methods in
  Natural Language Processing}, pages 2769--2774, Brussels, Belgium.
  Association for Computational Linguistics.

\bibitem[{Lowe et~al.(2015)Lowe, Pow, Serban, and Pineau}]{ubuntu}
Ryan Lowe, Nissan Pow, Iulian Serban, and Joelle Pineau. 2015.
\newblock \href {https://doi.org/10.18653/v1/W15-4640} {The {U}buntu dialogue
  corpus: A large dataset for research in unstructured multi-turn dialogue
  systems}.
\newblock In \emph{Proceedings of the 16th Annual Meeting of the Special
  Interest Group on Discourse and Dialogue}, pages 285--294, Prague, Czech
  Republic. Association for Computational Linguistics.

\bibitem[{Magerko et~al.(2009)Magerko, Manzoul, Riedl, Baumer, Fuller, Luther,
  and Pearce}]{magerkoemp}
Brian Magerko, Waleed Manzoul, Mark Riedl, Allan Baumer, Daniel Fuller, Kurt
  Luther, and Celia Pearce. 2009.
\newblock \href {https://doi.org/10.1145/1640233.1640253} {An empirical study
  of cognition and theatrical improvisation}.
\newblock In \emph{Proceedings of the Seventh ACM Conference on Creativity and
  Cognition}, page 117–126, New York, NY, USA. Association for Computing
  Machinery.

\bibitem[{Martin et~al.(2016)Martin, Harrison, and
  Riedl}]{riedl_improv_storytelling}
Lara~J. Martin, Brent Harrison, and Mark~O. Riedl. 2016.
\newblock Improvisational computational storytelling in open worlds.
\newblock In \emph{Interactive Storytelling}, pages 73--84, Cham. Springer
  International Publishing.

\bibitem[{McHugh(2016)}]{mchugh2016podcasting}
Siobhan McHugh. 2016.
\newblock How podcasting is changing the audio storytelling genre.
\newblock \emph{Radio Journal: International Studies in Broadcast \& Audio
  Media}, 14(1):65--82.

\bibitem[{Papineni et~al.(2002)Papineni, Roukos, Ward, and
  Zhu}]{papineni2002bleu}
Kishore Papineni, Salim Roukos, Todd Ward, and Wei-Jing Zhu. 2002.
\newblock \href {https://doi.org/10.3115/1073083.1073135} {{B}leu: {a} method
  for automatic evaluation of machine translation}.
\newblock In \emph{Proceedings of the 40th Annual Meeting of the Association
  for Computational Linguistics}, pages 311--318, Philadelphia, Pennsylvania,
  USA. Association for Computational Linguistics.

\bibitem[{P{\~o}ldvere et~al.(2017)P{\~o}ldvere, Johansson, and
  Paradis}]{poldvere2017london}
Nele P{\~o}ldvere, V~Johansson, and C~Paradis. 2017.
\newblock The {L}ondon-{L}und corpus 2: A new corpus of spoken {B}ritish
  {E}nglish in the making.
\newblock In \emph{theICAME 38 Conference, Prague, Czech Republic}.

\bibitem[{Radford et~al.(2019)Radford, Wu, Child, Luan, Amodei, and
  Sutskever}]{radford2019language}
Alec Radford, Jeffrey Wu, Rewon Child, David Luan, Dario Amodei, and Ilya
  Sutskever. 2019.
\newblock Language models are unsupervised multitask learners.
\newblock Technical report, OpenAI.

\bibitem[{Serban et~al.(2015)Serban, Sordoni, Bengio, Courville, and
  Pineau}]{serban_hierarchical_generative_nn}
Iulian Serban, Alessandro Sordoni, Yoshua Bengio, Aaron~C. Courville, and
  Joelle Pineau. 2015.
\newblock Hierarchical neural network generative models for movie dialogues.
\newblock \emph{ArXiv}, abs/1507.04808.

\bibitem[{Serban et~al.(2017)Serban, Sordoni, Lowe, Charlin, Pineau, Courville,
  and Bengio}]{serban2017hierarchical}
Iulian Serban, Alessandro Sordoni, Ryan Lowe, Laurent Charlin, Joelle Pineau,
  Aaron Courville, and Yoshua Bengio. 2017.
\newblock A hierarchical latent variable encoder-decoder model for generating
  dialogues.
\newblock In \emph{Thirty-First AAAI Conference on Artificial Intelligence}.

\bibitem[{Shao et~al.(2017)Shao, Gouws, Britz, Goldie, Strope, and
  Kurzweil}]{shao2017}
Louis Shao, Stephan Gouws, Denny Britz, Anna Goldie, Brian Strope, and Ray
  Kurzweil. 2017.
\newblock Generating long and diverse responses with neural conversation
  models.
\newblock \emph{arXiv preprint arXiv:1701.03185}.

\bibitem[{Sordoni et~al.(2015)Sordoni, Galley, Auli, Brockett, Ji, Mitchell,
  Nie, Gao, and Dolan}]{sordoni_context_sensitive_generation}
Alessandro Sordoni, Michel Galley, Michael Auli, Chris Brockett, Yangfeng Ji,
  Margaret Mitchell, Jian-Yun Nie, Jianfeng Gao, and Bill Dolan. 2015.
\newblock \href {https://doi.org/10.3115/v1/N15-1020} {A neural network
  approach to context-sensitive generation of conversational responses}.
\newblock In \emph{Proceedings of the 2015 Conference of the North {A}merican
  Chapter of the Association for Computational Linguistics: Human Language
  Technologies}, pages 196--205, Denver, Colorado. Association for
  Computational Linguistics.

\bibitem[{Spolin et~al.(1986)Spolin, Morey, and Brandt}]{spolin1986theater}
Viola Spolin, Arthur Morey, and Mary~Ann Brandt. 1986.
\newblock \href {https://books.google.com/books?id=\_Cp-xvnCEgIC}
  {\emph{Theater Games for the Classroom: A Teacher's Handbook}}.
\newblock Northwestern University Press.

\bibitem[{Traum(1999)}]{traum1999computational}
David~R Traum. 1999.
\newblock Computational models of grounding in collaborative systems.
\newblock In \emph{Psychological Models of Communication in Collaborative
  Systems-Papers from the AAAI Fall Symposium}, pages 124--131.

\bibitem[{Winston and Magerko(2017)}]{winston_magerko}
Lauren Winston and Brian Magerko. 2017.
\newblock Turn-taking with improvisational co-creative agents.
\newblock In \emph{Thirteenth Artificial Intelligence and Interactive Digital
  Entertainment Conference}.

\bibitem[{Wolf et~al.(2019{\natexlab{a}})Wolf, Debut, Sanh, Chaumond, Delangue,
  Moi, Cistac, Rault, Louf, Funtowicz, and Brew}]{Wolf2019HuggingFacesTS}
Thomas Wolf, Lysandre Debut, Victor Sanh, Julien Chaumond, Clement Delangue,
  Anthony Moi, Pierric Cistac, Tim Rault, R'emi Louf, Morgan Funtowicz, and
  Jamie Brew. 2019{\natexlab{a}}.
\newblock Huggingface's transformers: State-of-the-art natural language
  processing.
\newblock \emph{ArXiv}, abs/1910.03771.

\bibitem[{Wolf et~al.(2019{\natexlab{b}})Wolf, Sanh, Chaumond, and
  Delangue}]{transfer-learning-conversation}
Thomas Wolf, Victor Sanh, Julien Chaumond, and Clement Delangue.
  2019{\natexlab{b}}.
\newblock Transfertransfo: {A} transfer learning approach for neural network
  based conversational agents.
\newblock \emph{arXiv preprint arXiv:1901.08149}.

\bibitem[{Xu et~al.(2017)Xu, Liu, Wang, Sun, Wang, Wang, and
  Qi}]{xu-etal-2017-neural}
Zhen Xu, Bingquan Liu, Baoxun Wang, Chengjie Sun, Xiaolong Wang, Zhuoran Wang,
  and Chao Qi. 2017.
\newblock \href {https://doi.org/10.18653/v1/D17-1065} {Neural response
  generation via {GAN} with an approximate embedding layer}.
\newblock In \emph{Proceedings of the 2017 Conference on Empirical Methods in
  Natural Language Processing}, pages 617--626, Copenhagen, Denmark.
  Association for Computational Linguistics.

\bibitem[{Yarowsky(1995)}]{yarowsky-1995-unsupervised}
David Yarowsky. 1995.
\newblock \href {https://doi.org/10.3115/981658.981684} {Unsupervised word
  sense disambiguation rivaling supervised methods}.
\newblock In \emph{33rd Annual Meeting of the Association for Computational
  Linguistics}, pages 189--196, Cambridge, Massachusetts, USA. Association for
  Computational Linguistics.

\bibitem[{Zhang et~al.(2018)Zhang, Dinan, Urbanek, Szlam, Kiela, and
  Weston}]{convai}
Saizheng Zhang, Emily Dinan, Jack Urbanek, Arthur Szlam, Douwe Kiela, and Jason
  Weston. 2018.
\newblock \href {https://doi.org/10.18653/v1/P18-1205} {Personalizing dialogue
  agents: {I} have a dog, do you have pets too?}
\newblock In \emph{Proceedings of the 56th Annual Meeting of the Association
  for Computational Linguistics (Volume 1: Long Papers)}, pages 2204--2213,
  Melbourne, Australia. Association for Computational Linguistics.

\bibitem[{Zhao et~al.(2017)Zhao, Zhao, and Eskenazi}]{zhao2017diversity_vae}
Tiancheng Zhao, Ran Zhao, and Maxine Eskenazi. 2017.
\newblock \href {https://doi.org/10.18653/v1/P17-1061} {Learning
  discourse-level diversity for neural dialog models using conditional
  variational autoencoders}.
\newblock In \emph{Proceedings of the 55th Annual Meeting of the Association
  for Computational Linguistics (Volume 1: Long Papers)}, pages 654--664,
  Vancouver, Canada. Association for Computational Linguistics.

\end{thebibliography}
